\pdfoutput=1
\documentclass[11pt]{article}

\usepackage[final]{main}

\usepackage{natbib}
\usepackage{times}
\usepackage{latexsym}
\usepackage{graphicx}
\usepackage{placeins}
\usepackage{float} 
\usepackage{booktabs} 
\usepackage{tabularx}
\usepackage{amssymb}
\usepackage{amsmath}
\usepackage{caption}
\usepackage{multirow}
\usepackage[T1]{fontenc}
\usepackage[utf8]{inputenc}

\usepackage{microtype}

\usepackage{inconsolata}

\title{Layer-Aware Embedding Fusion for LLMs in Text Classifications}

\author{Jiho Gwak \quad Yuchul Jung \\
  Department of Computer Engineering,\\ Kumoh National Institute of Technology \\
  Gumi, Korea\\
  \texttt{wlgh6709@kumoh.ac.kr} \quad \texttt{jyc@kumoh.ac.kr}
}

\begin{document}
\maketitle
\begin{abstract}
Embedding fusion has emerged as an effective approach for enhancing performance across various NLP tasks. However, systematic guidelines for selecting optimal layers and developing effective fusion strategies for the integration of LLMs remain underexplored. In this study, we propose a layer-aware embedding selection method and investigate how to quantitatively evaluate different layers to identify the most important ones for downstream NLP tasks, showing that the critical layers vary depending on the dataset. We also explore how combining embeddings from multiple LLMs, without requiring model fine-tuning, can improve performance. Experiments on four English text classification datasets (SST-2, MR, R8, and R52) demonstrate that different layers in LLMs exhibit varying degrees of representational strength for classification, and that combining embeddings from different models can enhance performance if the models exhibit complementary characteristics. Additionally, we discuss resources overhead (memory and inference time) to provide a balanced perspective on the real world feasibility of embedding fusion. Future work will explore multilingual and domain specific datasets, as well as techniques for automating layer selection, to improve both performance and scalability.
\end{abstract}

\section{Introduction}

With the recent advancements in large language models (LLMs), the representational capacity 
of decoder-based models \citep{brown2020languagemodelsfewshotlearners, touvron2023llamaopenefficientfoundation, touvron2023llama2openfoundation} has attracted 
considerable attention in NLP downstream tasks \citep{zhang-etal-2022-prompt-based, sun2023textclassificationlargelanguage}. 
Despite their impressive zero or few-shot performance, these models are primarily trained 
for next token prediction, leading to layer-wise differences in how semantic and contextual 
information is encoded. Traditional usage often relies on final layer embeddings 
\citep{brown2020languagemodelsfewshotlearners}, yet previous studies in encoder-based architectures 
\citep{devlin2019bertpretrainingdeepbidirectional} have hinted that intermediate layers may yield richer representations 
for classification \citep{zhang2024investigatinglayerimportancelarge}.

Moreover, as diverse pretrained models including generative LLMs and specialized embedding 
models continue to proliferate \citep{lee2025nvembedimprovedtechniquestraining, wang2024textembeddingsweaklysupervisedcontrastive}, model fusion has emerged 
as a practical approach to leverage complementary knowledge. However, systematic guidelines 
for (1) which layer to select from a decoder-based LLM for a particular task, and (2) how to 
efficiently fuse embeddings across multiple LLMs with minimal computational overhead, remain limited.

To address these challenges, we present a series of contributions aimed at improving 
layer selection and model combination in LLM-based embeddings.

\noindent\textbf{Layer-Aware Selection}\hspace{0.5em} 
We present a method that empirically demonstrates the importance of layer selection 
through quantitative experiments, showing that specific layers are crucial for text 
classification. The results provide both empirical and partially theoretical insights 
into why certain mid/late layers outperform the final layer in decoder-based LLMs.

\noindent\textbf{Layer-Aware Embedding Fusion}\hspace{0.5em} 
We demonstrate the fusion of embeddings from multiple LLM models without fine-tuning, 
showing how combining different models improves performance across various NLP tasks. 
By considering the layers of the LLM models, we achieve optimal performance through 
embedding fusion, proving that specific layers play a crucial role in determining 
the most effective combination.

\noindent\textbf{Stabilizing Performance through Multi-Model Fusion}\hspace{0.5em} 
Through experiments combining more than three models, we show that classification performance 
becomes more stable as more models are integrated, providing empirical evidence of the benefits 
of multi-model fusion for improving task-specific accuracy.

\section{Related Work}
\subsection{Text Classification and Pretrained Language Models}
Text classification has long been a central research topic in the field of natural language processing (NLP). 
In the early stages, statistical representation techniques such as bag-of-words, term frequency-inverse document frequency (TF-IDF), and $n$ grams, along with traditional machine learning models such as support vector machines (SVM)~\citep{pang2002thumbsupsentimentclassification} and logistic regression~\citep{inproceedings}, dominated the field.

With the rapid advancement of deep learning, particularly the introduction of pretraining language models, the paradigm of text classification has undergone a significant change. 
Representative models such as BERT (Bidirectional Encoder Representations from Transformers)~\citep{devlin2019bertpretrainingdeepbidirectional} and RoBERTa (Robustly Optimized BERT Pre-Training Approach)~\citep{liu2019robertarobustlyoptimizedbert} significantly improved the ability to capture contextual information in text through bidirectional encoder architectures and large-scale pre-training corpora.
By fine-tuning these models on downstream tasks, they have been shown to achieve substantially higher classification performance compared to traditional approaches~\citep{youngmin2024rolemodelarchitecturescale}.

\subsection{Extended Applications of LLM}

The evolution of Large Language Models (LLMs) based on decoder architectures has enabled zero-shot~\citep{kojima2023largelanguagemodelszeroshot} and few-shot~\citep{zhang-etal-2022-prompt-based} learning for downstream tasks such as text classification, spurring research on prompt-based approaches such as chain-of-thought (CoT)~\citep{wei2023chainofthoughtpromptingelicitsreasoning} and CARP~\citep{sun2023textclassificationlargelanguage}. 
Numerous studies have reported on the performance of LLMs in classification tasks, including models such as GPT-3~\citep{brown2020languagemodelsfewshotlearners} and the LLaMA series~\citep{touvron2023llamaopenefficientfoundation, touvron2023llama2openfoundation}, as well as empirical studies analyzing their behavior~\citep{sarkar-etal-2023-zero, gretz-etal-2023-zero}. 
However, these methods exhibit considerable performance variability depending on prompt design~\citep{cao2024worstpromptperformancelarge, he2024doespromptformattingimpact}.

Meanwhile, there is growing interest in using LLMs not only as generative models but also as providers of high-quality embeddings~\citep{tao2024llmseffectiveembeddingmodels}. 
Recent research suggests that relatively lightweight LLMs (e.g. up to 7B parameters) can produce strong embedding quality with efficient computational resources~\citep{wang2024textembeddingsweaklysupervisedcontrastive, lee2025nvembedimprovedtechniquestraining}. 
Furthermore, several studies have emphasized that different embedding layers within a model can produce optimal representations for tasks, with a particular focus on the importance of layer selection~\citep{zhang2024investigatinglayerimportancelarge}. 
These findings highlight the role of intermediate representations in understanding the encoding behavior of transformer models in various applications~\citep{skean2024doesrepresentationmatterexploring}.

\subsection{Embedding Fusion}

As diverse pretrained models, including large language models (LLMs), continue to emerge, there has been increasing interest in combining embeddings extracted from multiple models~\citep{shinnou-etal-2018-domain, blandfort2019fusionstrategieslearninguser}. 
Previous studies have reported performance improvements on tasks such as text classification and sentiment analysis through embedding fusion.

For example, LLMEmbed~\citep{liu2024llmembedrethinkinglightweightllms} demonstrates that combining embeddings from LLaMA2~\citep{touvron2023llama2openfoundation} with those from BERT and RoBERTa can effectively leverage the distinctive representational characteristics of each model. 
Moreover, a variety of fusion strategies have been proposed not only in NLP, but also in other domains for instance, QUARC~\citep{kumar2020quarcquaternionmultimodalfusion} applies quaternion-based operations, while FuseMoE~\citep{han2024fusemoemixtureofexpertstransformersfleximodal} adopts a mixture-of-experts (MoE) architecture.

However, significant performance differences arise depending on the fusion method employed, and not all combinations lead to consistent improvements~\citep{Ko2024.08.24.609531}. 
Furthermore, since each embedding layer possesses a different representational capacity, understanding the layer-wise characteristics is critical for effective fusion~\citep{kaushik2024enhancingauthorshipattributionembedding}.

\begin{figure*}
    \centering
    \includegraphics[width=1\textwidth]{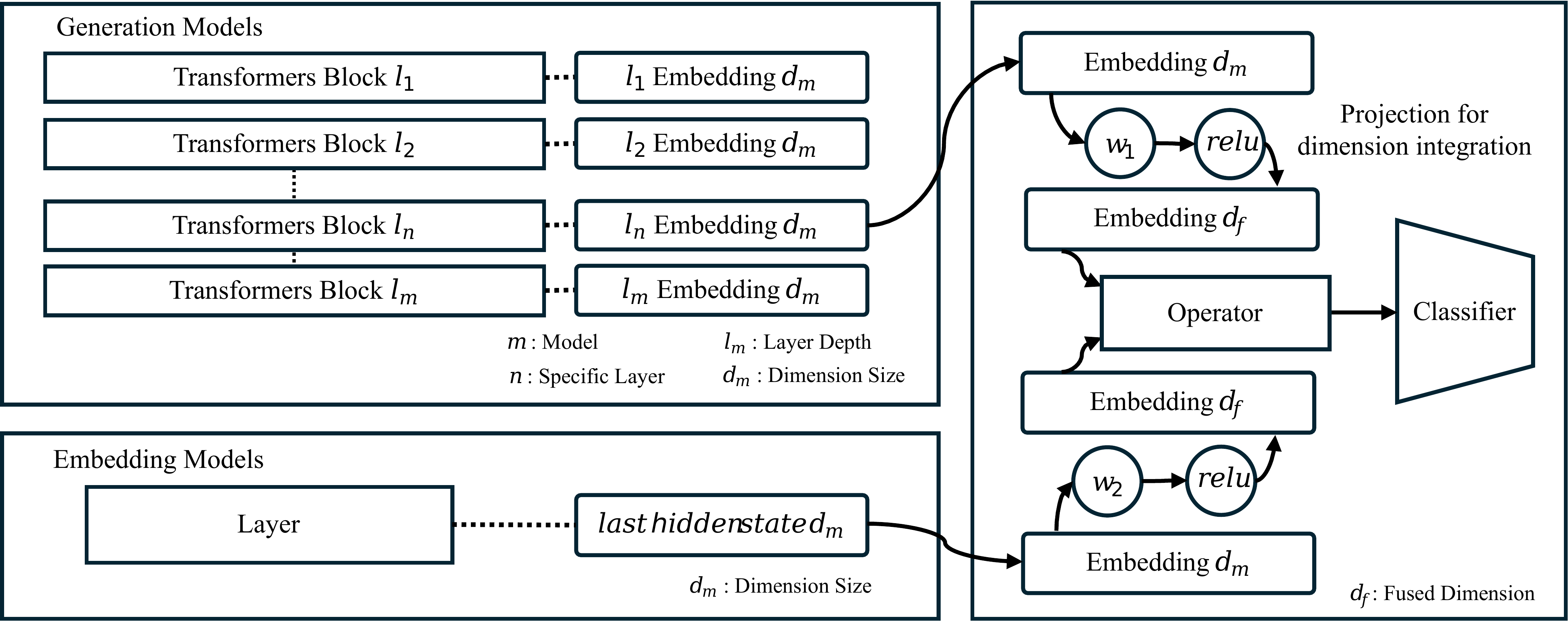}
    \caption{The embeddings extracted from the LLM are mapped to a unified dimension using a linear projection, after which various fusion techniques are applied. The selected layer n from the generation models represents the most informative layer for classification The two embeddings, normalized to dimension $d_{M(M=1024)}$, are then combined using a specific fusion strategy to generate a new representation. Finally, this fused embedding is fed into the classifier head, which is trained to optimize classification performance.}
    \label{fig:architecture}
\end{figure*}

\section{Methodology}
In this study, we enhance text classification performance using embeddings from various LLMs through three primary perspectives. First, we quantitatively examine the performance differences when using embeddings from a single layer versus combining embeddings from multiple layers in decoder-based LLMs.

Second, we compare and analyze strategies to enhance performance by combining embeddings from different LLMs. Lastly, we explore the potential for further performance improvement when combining embeddings from three or more LLMs. The experiments utilize generation models (with up to 7B parameters) and embedding models. The following table summarizes the key specifications of the models used in the study.

\setlength{\textfloatsep}{2em}
\begin{table}[t]
\centering
\caption{Summary of the key specifications of the generation models and embedding models utilized in the experiments, including model name, embedding dimension (Dim), and parameter size.}
\begin{tabularx}{\linewidth}{>{\raggedright\arraybackslash}X c c}
\toprule
\textbf{Model} & \textbf{Dim} & \textbf{Parameters} \\
\midrule
\texttt{LLaMA2}\citep{touvron2023llamaopenefficientfoundation, touvron2023llama2openfoundation} & 4096 & 6.92B \\
\texttt{Qwen2.5}\citep{qwen2025qwen25technicalreport}                                & 3584  & 7.62B \\
\texttt{Falcon 3}\citep{almazrouei2023falconseriesopenlanguage}                         & 3072  & 6.98B \\
\texttt{Mistral}\citep{jiang2023mistral7b}                               & 4096  & 6.92B \\
\texttt{Gemma 2}\citep{gemmateam2024gemma2improvingopen}                             & 2304  & 2B \\
\texttt{NV-Embed-v2}\citep{lee2025nvembedimprovedtechniquestraining}                             & 4096  & 7.10B \\
\texttt{e5-large-v2}\citep{wang2024textembeddingsweaklysupervisedcontrastive}                            & 1024  & 0.335B \\
\bottomrule
\end{tabularx}
\vspace{-1em}
\label{tab:your_table}
\end{table}

\subsection{Strategies for Using Embeddings from Generation Models}
\noindent\textbf{Last Layer vs. Layer wise Performance}\hspace{0.5em} First, it is known that the embedding representation capacity of decoder-based LLMs generally increases as the layers deepen. To verify this, this study extracts the hidden states of all layers and generates embeddings for each layer to measure text classification performance. Through this process, we aim to identify the optimal layers that yield the best classification performance.

\noindent\textbf{Multi-Layer Representation Aggregation}\hspace{0.5em} Second, we investigate how performance variations when combining embeddings from the last 1 to 10 layers. Fusion methods such as averaging, max, and min are applied to integrate the embeddings. For instance, if embeddings are extracted from the last three layers, we average them to create a single embedding and input it into the classification model. This is expected to provide more stable and consistent performance compared to a single layer, while methods like max or min fusion may cause performance fluctuations due to inherent biases in element selection.

\noindent\textbf{Single Layer vs Multiple Layers}\hspace{0.5em} Finally, based on the results of the previous two experiments, we compare the use of a single layer versus combining multiple layers. This comparison quantitatively evaluates whether combining multiple layers significantly improves classification performance or if a single optimal layer can achieve sufficient performance, considering the additional computational cost and memory usage.

\subsection{Integrating Embeddings from Various Models}
\paragraph{Linear Projection for Embedding Dimension Integration}
Embeddings extracted from multiple models may have different dimensions, so we apply a linear projection to unify them before combining.
For example, to project an embedding $E \in \mathbb{R}^{d_1}$ to a target dimension $d_2$, we use learnable parameters: a weight matrix $W \in \mathbb{R}^{d_2 \times d_1}$ and a bias vector $b \in \mathbb{R}^{d_2}$.
To incorporate nonlinearity, we apply a transformation such as: To incorporate nonlinearity, we apply a transformation such as $f(E) = \mathrm{ReLU}(W E + b)$. These projected embeddings are then integrated using various fusion techniques, allowing them to coexist within a unified dimensional space.
In this study, we perform the projection onto the smaller dimension and apply the ReLU activation function~\citep{agarap2019deeplearningusingrectified} to introduce nonlinear characteristics into the embeddings.

\paragraph{Fusion Methods}
A key focus of this study is to enhance representational capabilities of various models by combining embeddings extracted from various LLMs. To ensure the validity of the experiments across multiple models, embeddings from the last layers of each model are extracted and fused in various ways. The specific fusion techniques are as follows:

\begin{itemize}
    \item \textbf{Concatenation:} Directly concatenating the embedding vectors $E$ along the matrix dimension to form a single embedding:
    \begin{equation}
    E' = [E_1 \| \| E_2 \| \| \dots \| \| E_n]
    \end{equation}

    \item \textbf{Sum:} After aligning the dimensions of the embeddings, we form a new embedding by element-wise addition:
    \begin{equation}
    E' = \sum_{i=1}^{n} f(E_i)
    \end{equation}

    \item \textbf{Multiplication:} After aligning dimensions, embeddings are transformed into 2D arrays and combined via matrix multiplication.
    \begin{align}
    E_1 &\in \mathbb{R}^{d} \rightarrow \textit{reshape} \rightarrow E_1' \in \mathbb{R}^{32 \times 32} \notag \\
    E_2 &\in \mathbb{R}^{d} \rightarrow \textit{reshape} \rightarrow E_2' \in \mathbb{R}^{32 \times 32} \\
    E'  &= E_1' \cdot E_2' \notag \\
    E'' &= \textit{flatten}(E') \in \mathbb{R}^{1024} \label{eq:multiply_fusion} \notag
    \end{align}
    
    \item \textbf{Hadamard (Element-wise Product):} Embeddings are combined by multiplying corresponding elements at the same position across models.

    \item \textbf{Quaternion Fusion}~\citep{kumar2020quarcquaternionmultimodalfusion}: Embeddings are treated as quaternion-valued vectors and fused using quaternion operations, preserving multidimensional inter-model relationships.

    \item \textbf{Mixture-of-Experts Fusion}~\citep{han2024fusemoemixtureofexpertstransformersfleximodal}: Embeddings from each model are processed through different expert modules, and the final representation is generated via weighted sum across experts.

    \item \textbf{All Methods:} All fusion methods above are applied simultaneously and sequentially to observe how combinations influence performance.

    \item \textbf{Residually Enhanced Fusion}~\citep{gardias2020enhancedresidualnetworkscontextbased}: The newly generated embedding is combined with the original one via residual connections to incorporate additional information while preserving the original expressiveness.
\end{itemize}

\subsection{Fusion of Three or More LLMs}
While the previous sections focused on combining embeddings from two models, this section investigates the performance impact when combining embeddings from three or more LLMs.

The motivation behind combining multiple models is to leverage their complementary strengths, potentially maximizing performance improvement. The key objectives of this experiment are twofold: first, to determine whether combining embeddings from three or more models leads to performance improvement. second, to assess whether performance variance decreases as more models are combined, thus improving stability. 

The fusion method used for combining embeddings in this experiment is primarily concatenation, and all possible combinations are tested. Performance is evaluated using the same method described in section 3.2.

\subsection{Enhancing Fusion by Selecting Meaningful Layers}
Through the experiments in section 3.1, we confirmed that the optimal embedding layer may vary depending on the dataset. In sections 3.2 and 3.3, we analyzed the effects of different embedding fusion techniques and combining multiple models. Building on these findings, this section examines whether selecting specific layers from particular models can optimize embedding performance.

By incorporating the optimal layer information derived from section 3.1 for each dataset, we compare performance between using a single layer and combining multiple layers. Additionally, we apply the optimal layer selection strategy from the analysis of model fusion techniques in sections 3.2 and 3.3 and quantitatively assess whether the combined embeddings contribute to performance improvement. This analysis aims to investigate the impact of combining embeddings based on the optimal layers for each dataset and model, Impacts the stability and consistency of classification performance.

\begin{table*}[t]
\centering
\caption{This table presents the accuracy for the three datasets. l represents the optimal layer for each model. Models without l in the table either achieve the highest accuracy at the last layer or correspond to embedding models. Bolded values indicate the best performance for each dataset.}
\begin{tabularx}{\textwidth}{p{1.5cm} p{2.5cm} >{\centering\arraybackslash}p{3.0cm} c c c c}
\toprule
\multicolumn{7}{l}{\textbf{Previous Work}} \\
\midrule
\multicolumn{2}{l}{\textbf{Method}} & \textbf{Backbone} & \textbf{SST2} & \textbf{MR} & \textbf{R8} & \textbf{R52} \\
\midrule
\multicolumn{2}{l}{{CARP~\citep{sun2023textclassificationlargelanguage}}} & LLaMA2 7B & 0.8842 & 0.8494 & 0.9676 & 0.7305 \\
\multicolumn{2}{l}{{CARP~\citep{sun2023textclassificationlargelanguage}}} & LLaMA2 7B & 0.9569 & 0.9074 & 0.9783 & 0.9627 \\
\multicolumn{2}{l}{{LLMEmbed~\citep{lee2025nvembedimprovedtechniquestraining}}} & LLaMA2 7B & 0.9576 & 0.9549 & 0.9822 & 0.9568 \\
\midrule

\multicolumn{7}{l}{\textbf{Fusion Method}} \\
\midrule
\multicolumn{2}{l}{\textbf{Embedding$_{\text{model,layer}}$}}  & \textbf{Fusion Method} & \textbf{SST2} & \textbf{MR} & \textbf{R8} & \textbf{R52} \\
\midrule
\multirow{12}{*}{\shortstack{Specific\\Layer}}
& $E_{\text{ll}(\text{LLaMA2}, L{=}32)}$      & -- & 0.9518 & 0.9586 & 0.9735 & 0.9381 \\
& {\scriptsize \textbf{Best }} $E_{\text{ll},L{=}20}$            & & 0.9522 & {0.9629} & {0.9794} & {0.9416} \\
& $E_{\text{mi}(\text{Mistral}, L{=}32)}$     & -- & 0.9232 & 0.9539 & 0.9639 & 0.9042 \\
& {\scriptsize \textbf{Best }} $E_{\text{mi},L{=}25}$            & & --     & --     & --     & {0.9136} \\
& $E_{\text{fa}(\text{Falcon3}, L{=}28)}$     & -- & 0.9220 & 0.9552 & 0.9657 & 0.8314 \\
& {\scriptsize \textbf{Best }} $E_{\text{fa},L{=}21}$            & & {0.9369} & {0.9589} & {0.9694} & {0.8376} \\
& $E_{\text{qw}(\text{Qwen}, L{=}28)}$        & -- & 0.9335 & 0.9589 & 0.9753 & 0.9412 \\
& {\scriptsize \textbf{Best }} $E_{\text{qw},L{=}20}$            & & {0.9484} & {0.9632} & {0.9772} & {0.9451} \\
& $E_{\text{ge}(\text{Gemma2}, L{=}26)}$      & -- & 0.9209 & 0.9564 & 0.9753 & 0.8742 \\
& {\scriptsize \textbf{Best }} $E_{\text{ge},L{=}19}$            & & {0.9278} & {0.9601} & {0.9781} & -- \\
& $E_{\text{nv}(\text{NV-Embed-v2})}$         & -- & 0.9564 & 0.9699 & 0.9808 & 0.9595 \\
& $E_{\text{e5}(\text{e5\_large\_v2})}$       & -- & 0.9461 & 0.9545 & 0.9785 & 0.9583 \\

\midrule
\multirow{4}{*}{\shortstack{Two \\ Models}}
& $E_{\text{e5}}, E_{\text{nv}}$ & $\text{Quaternion}(R)$ & \textbf{0.9644} & 0.9709 & 0.9840 & 0.9595 \\
& $E_{\text{nv}}, E_{\text{ge},L{=}l}$ & $\text{Hadamard}(R)$ & {0.9564}\textsubscript{l=19} & {0.9709}\textsubscript{l=20} & \textbf{0.9849}\textsubscript{l=23} & {0.9451}\textsubscript{l=23} \\
& $E_{\text{ll},L{=}l}, E_{\text{nv}}$ & $\text{Quaternion}(R)$ & \textbf{0.9644}\textsubscript{l=20} & {0.9702}\textsubscript{l=20} & {0.9826}\textsubscript{l=27} & {0.9591}\textsubscript{l=28} \\
& $E_{\text{qw},L{=}l}, E_{\text{nv}}$ & $\text{Multiplication}(R)$ & {0.9587}\textsubscript{l=20} & \textbf{0.9721}\textsubscript{l=20} & {0.9836}\textsubscript{l=27} & {0.9626}\textsubscript{l=27} \\
\midrule

\multirow{2}{*}{\shortstack{Multi\\Models}} 
& $E_{\text{ll}}, E_{\text{qw}}, E_{\text{nv}}, E_{\text{e5}}$ & $\text{Concatenation}$ & 0.961 & 0.9709 & 0.9845 & \textbf{0.9638} \\
& $\text{ALL}(E)$ & $\text{Sum}(R)$ & 0.961 & 0.9719 & 0.9822 & 0.9611 \\
\bottomrule
\end{tabularx}
\label{tab:previous_fusion_combined}
\end{table*}

\section{Experiment}
Text classification performance using the combined embeddings is primarily evaluated based on accuracy. Given the extensive nature of the experimental results, we present a summary of the most significant or representative results in table format. 

To evaluate text classification performance, this study utilized SST-2~\citep{socher-etal-2013-recursive} MR~\citep{maas-etal-2011-learning}, and R8 datasets. SST-2 and MR are binary sentiment classification datasets based on movie reviews, with 67,349/872 (train/test) and 40,000/10,000 samples, respectively. R8, derived from Reuters-21578, is a document classification dataset with 5,485/2,189 samples across 8 categories. R52, also derived from Reuters-21578, is a larger variant consisting of 6,532 training and 2,568 test samples distributed across 52 categories.
In this study, experiments were conducted in an environment equipped with two NVIDIA RTX 4090 GPUs (24GB each). To perform text classification using the fused embedding vectors, a multi-layer perceptron (MLP)-based classifier was employed. During training, the batch size was set to 100, the learning rate to 1e-4, the optimizer to Adam, and the number of epochs to 120.

This chapter systematically analyzes the performance variations observed when using embeddings extracted from decoder-based LLMs for text classification. The analysis focuses on two main aspects: (1) performance across specific layers, and (2) the impact of fusing embeddings from different models.

\begin{figure}
    \centering
    \includegraphics[width=\linewidth]{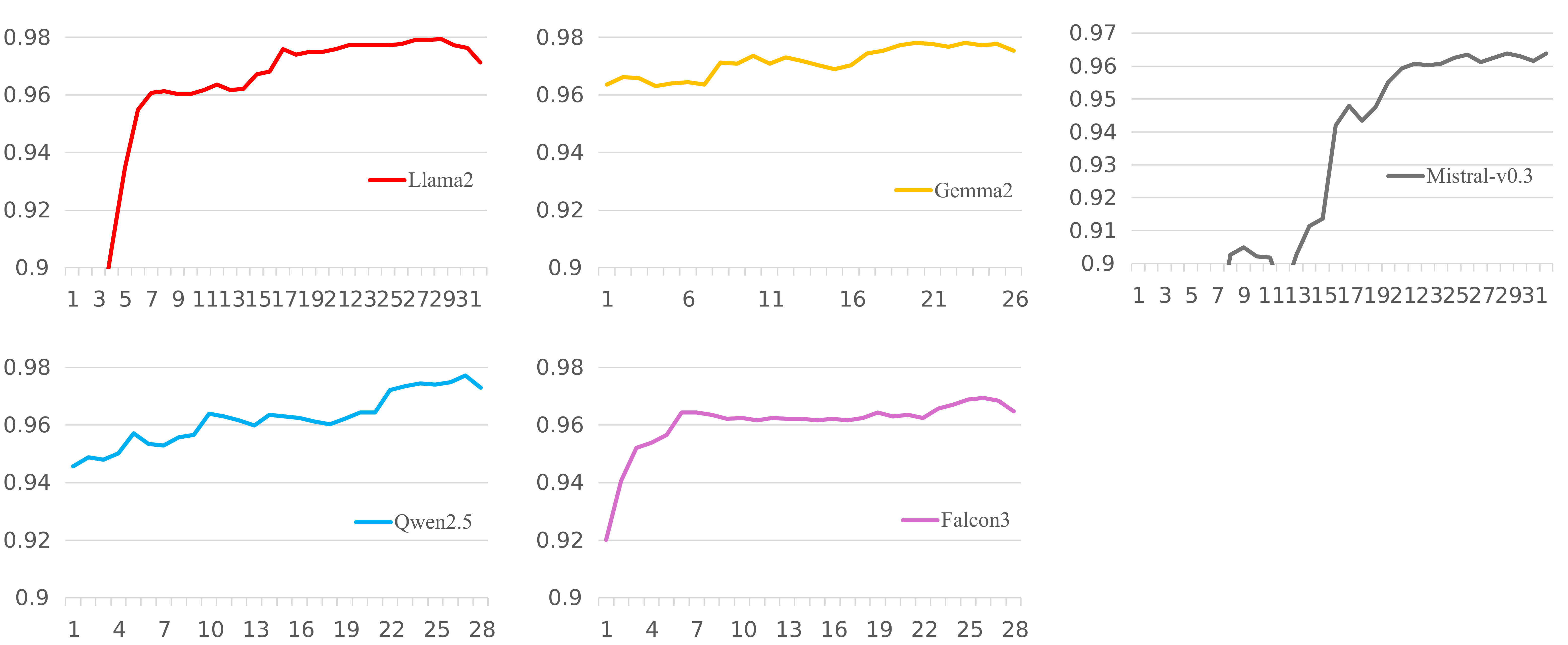}
    \caption{Comparison of Single-Layer Embedding Classification Performance in Decoder-Based LLMs}
    \label{fig:enter-label}
\end{figure}

\begin{figure}
    \centering
    \includegraphics[width=\linewidth]{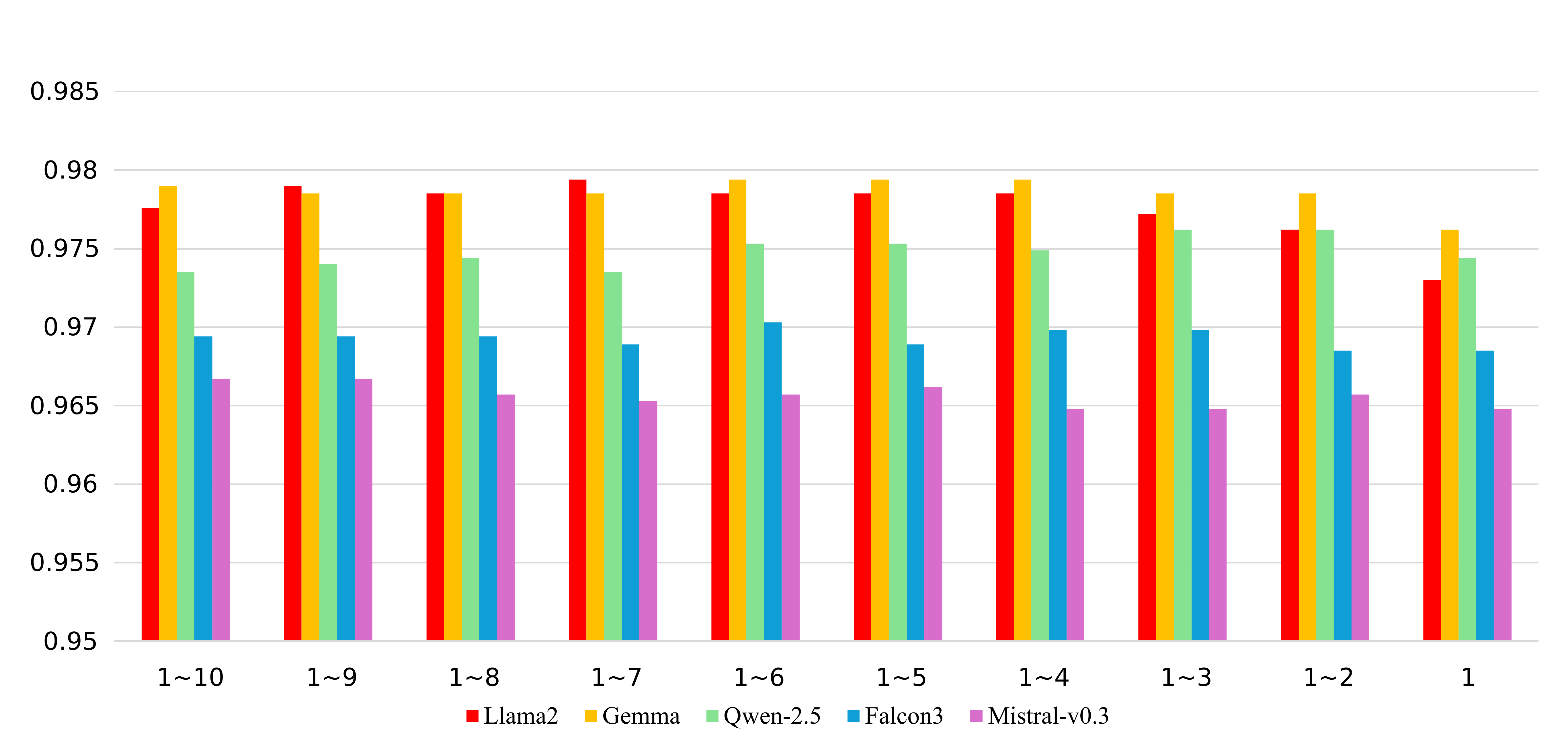}
    \caption{Comparison of Averaged Layer Embedding Fusion in Decoder-Based LLMs}
    \label{fig:enter-label2}
\end{figure}

\subsection{Layer Selection Strategies for Embeddings}
\noindent\textbf{Single Layer}\hspace{0.5em} As shown in Figure 2, classification performance generally increases toward the upper layers but tends to drop slightly at the final layer. This pattern suggests that the penultimate or nearby layers yield more discriminative and stable representations for classification tasks, with an average performance difference of approximately 0.04.

\noindent\textbf{Multiple Layer}\hspace{0.5em} As shown in Figure 3, the averaging-based fusion method generally outperformed the embedding from the final layer, although its performance remained lower than that of a well performing specific layer. Fusion methods based on maximum or minimum values exhibited performance comparable to that of the final layer. While combining multiple layers can lead to more stable representations, it does not necessarily yield better results than appropriately selecting a single representative layer. Furthermore, such approaches introduce additional memory usage and computational overhead.
\subsection{Layer Selection in Single Models for Classification}
According to section 4.1, we conducted further experiments using a single layer that demonstrated high performance. Instead of utilizing the final layer, we selected a specific intermediate layer, which showed an average performance improvement of approximately +0.4. However, this improvement was not consistent across all models. The optimal layer varied depending on the dataset: for SST-2 and MR, layers closer to the middle tended to yield better performance, whereas for R8 and R52, later layers performed better. The specific layers used in each case are presented in the “Specific Layer” row of Table 2.

\subsection{Layer-Aware Fusion of Embeddings from Two Models} Across datasets, the best performance was achieved by selecting optimal layers from each model, rather than using final-layer representations. Compared to NV\_embed~\citep{lee2025nvembedimprovedtechniquestraining}, the strongest-performing single model, fusion methods showed improvements of +0.080 on SST-2, +0.022 on MR, +0.041 on R8, and +0.043 on R52.

SST-2, MR, and R8 achieved the highest scores through layer-wise selective fusion, the corresponding results are presented in the Two Model row of Table 2. Notably, combining two individually strong models did not always result in superior performance, highlighting that model complementarity is more critical than standalone strength in fusion-based approaches.
\subsection{Combining Embeddings from Multiple Models}
Combining embeddings from multiple models generally resulted in improved performance. In nearly all cases, multi-model fusion outperformed single-model baselines. The highest performance on the R52 dataset was achieved by combining embeddings from four models, with an improvement of +0.0043.

For the other datasets, however, the highest performance was achieved by fusing embeddings from two models with appropriately selected layers. While the optimal number of models varied depending on the dataset, combining more models tended to improve classification performance in general. These results are presented in the “Multi-Model” row of Table 2.

\section{Results Analysis}
\subsection{Decline in Performance at the Last Layers}
The performance degradation observed in the final layers of decoder-based LLMs is closely related to their training objective. These models are primarily trained for next token prediction, leading them to emphasize token level interactions and localized patterns rather than capturing a global semantic representation. As a result, the later layers, particularly the final layer, become increasingly specialized for generation tasks, focusing on syntactic structures and positional cues essential for predicting the next token~\citep{kaushik2024enhancingauthorshipattributionembedding}.
In contrast, classification tasks require a broader contextual understanding and inference of semantic relationships, making the middle-to-late layers more effective than the final layer in many cases. These intermediate layers capture richer semantic representations, which can be beneficial for tasks involving general meaning inference and contextual comprehension~\citep{skean2024doesrepresentationmatterexploring}.
Therefore, embeddings extracted from the middle-to-late layers are often more effective than those from the final layer for classification tasks.

\subsection{Effect of Embedding Fusion}
To enhance the performance of downstream tasks, embedding fusion combining embeddings from multiple layers has been widely explored. By integrating features from different layers, this approach enables models to generate richer contextual representations, resulting in more comprehensive and robust embeddings. Embedding fusion not only enhances semantic expressiveness but also mitigates noise, improving overall stability and generalization across tasks. Experimental results have demonstrated performance improvements through this technique.

\subsection{Complementarity Between Different Models}
When combining different models, the unique characteristics and biases of each model play a crucial role. Since each model is trained differently, integrating complementary information from various models can lead to performance improvements.

Complementarity arises when models capture different types of information. By combining models that encode distinct representational patterns, the resulting embeddings can provide a richer and more informative signal for downstream tasks. However, when models encode similar or overlapping information, fusion tends to be less effective, sometimes even degrading performance due to redundancy or conflicting representational features.

Therefore, model selection is critical when considering combination. Selecting models with complementary representational characteristics is more likely to enhance performance, making it essential to analyze and understand the strengths of each model before fusion.

\subsection{Efficiency Analysis}
Memory consumption increases linearly with the number of fused models due to the expansion of the combined embedding dimension. For example, fusing embeddings from two models (e.g., NV-embed and e5) resulted in a combined embedding size of 5120 dimensions, requiring approximately 1.3 GB of storage for SST-2. However, adding more models rapidly increased memory requirements: combining embeddings from five models (e.g., NV-embed, e5, LLaMA2, Qwen, and Mistral) resulted in a combined embedding size of 16,896 dimensions and required approximately 4.3 GB—over 4× more memory compared to the two model case.

Memory growth is primarily driven by the increase in the fused embedding size, as each additional model contributes its own feature dimensions, leading to a proportional increase in storage and computational costs. While more models provide additional features, the diminishing returns in performance highlight the importance of balancing the trade off between accuracy gains and memory efficiency. Techniques such as dimensionality reduction or learned projections can mitigate memory growth while preserving performance.

\section{Conclusion}
This study analyzed various strategies for improving text classification performance using embeddings from large language models (LLMs). The analysis compared the effectiveness of single-layer and multi-layer embeddings, and experi-mentally investigated the impact of combining embeddings from different LLMs. Based on these results, optimal strategies for embedding fusion were discussed.

Improving text classification performance with LLM embeddings requires more than simply applying a fusion strategy. Our findings show that embeddings from mid-to-late layers generally outperform those from the final layer, which tend to encode generation specific or position heavy signals. While averaging across multiple layers yields more stable performance, it also increases memory and computational costs. When per-dataset layer selection is infeasible, selectively averaging late layers or empirically identifying a single effective layer can offer a practical alternative.

In multi-model settings, combining embeddings from different LLMs yields modest performance gains, but only when the models encode complementary information. Redundant or similarly biased models provide limited benefit and may introduce overfitting or inefficiencies. Although combining more than two models can improve performance stability, it also significantly increases resource demands. Therefore, task specific and resource aware fusion strategies, grounded in a careful analysis of model and layer characteristics, are essential for designing scalable and effective text classification systems.

\section{Limitation}
This study presents promising strategies for enhancing text classification performance but is limited by its focus on widely used English datasets (SST-2, MR, R8, R52), leaving the effectiveness in multilingual or domain specific contexts (e.g., medical, legal) largely unverified. This narrow focus on general purpose English classification represents a significant constraint on the generaliz ability of our findings.

The embedding fusion strategies explored showed performance improvements across several conditions but not consistently across all combinations. The optimal layer or model selection may vary depending on dataset characteristics, requiring repeated experimentation to identify the most effective configuration. Additionally, while our dimension adaptive projection approach partially addresses computational resource costs, it does not fully eliminate the trial-and-error needed to discover an optimal fusion strategy.

\section{Future Works}
Future research should focus on developing methods for extracting task optimized embedding layers, while also evaluating the generalizability of fusion strategies across multilingual datasets and various domains (e.g., medical, legal, technical documents). This could involve analyzing how embeddings from different layers influence task specific performance and establishing frameworks for automatically identifying the most suitable layers for specific tasks or domains. Such approaches could ensure high performance across diverse languages and domains, while also reducing computational costs and resource requirements. These advancements are expected to significantly contribute to both the practical applicability and scalability of real-world natural language processing tasks.

\appendix

\section{Layer-wise experimental results on the decoder-based models}
\begin{table}[ht]
\caption{Layer-wise classification performance on the R8 dataset.}
\centering
\begin{tabular}{cccc}
\toprule
\textbf{Layer} & \textbf{Llama2} & \textbf{Qwen2.5} & \textbf{Gemma2} \\
\midrule
15 & 0.9671 & 0.9630 & 0.9689 \\
16 & 0.9680 & 0.9625 & 0.9703 \\
17 & 0.9758 & 0.9612 & 0.9744 \\
18 & 0.9740 & 0.9603 & 0.9753 \\
19 & 0.9749 & 0.9621 & 0.9772 \\
20 & 0.9749 & 0.9644 & \textbf{0.9781} \\
21 & 0.9758 & 0.9644 & 0.9776 \\
22 & 0.9772 & 0.9721 & 0.9767 \\
23 & 0.9772 & 0.9735 & 0.9781 \\
24 & 0.9772 & 0.9744 & 0.9772 \\
25 & 0.9772 & 0.9740 & 0.9776 \\
26 & 0.9776 & 0.9749 & 0.9753 \\
27 & 0.9790 & \textbf{0.9772} & - \\
28 & \textbf{0.9794} & 0.9730 & - \\
29 & 0.9790 & -      & - \\
30 & 0.9772 & -      & - \\
31 & 0.9762 & -      & - \\
32 & 0.9712 & -      & - \\
\bottomrule
\end{tabular}
\label{tab:model_layer_comparison}
\end{table}

The selected models represent the top three decoder-based generative language models, ranked by classification accuracy.
Table 3 and Table 4 present label-wise classification accuracy on the R8 and R52 datasets, respectively.
For both datasets, the highest performance was typically achieved at the final decoder layers, indicating that deeper representations carry more task-relevant semantic information. Although results for SST-2 and MR are not explicitly included, peak performance for those datasets was observed around the 20th to 21st layers.

\begin{table}[ht]
\caption{Layer-wise classification performance on the R52 dataset.}
\centering
\begin{tabular}{cccc}
\toprule
\textbf{Layer} & \textbf{Llama2} & \textbf{Qwen2.5} & \textbf{Mistral} \\
\midrule
15 & 0.9147 & 0.8766 & 0.8621 \\
16 & 0.9147 & 0.8773 & 0.8606 \\
17 & 0.9248 & 0.8731 & 0.8703 \\
18 & 0.9206 & 0.8727 & 0.8727 \\
19 & 0.9248 & 0.8707 & 0.8688 \\
20 & 0.9260 & 0.8769 & 0.8711 \\
21 & 0.9280 & 0.8746 & 0.8734 \\
22 & 0.9330 & 0.8863 & 0.8715 \\
23 & 0.9361 & 0.9015 & 0.8688 \\
24 & 0.9369 & 0.9210 & 0.9062 \\
25 & 0.9400 & 0.9260 & 0.9163 \\
26 & 0.9400 & 0.9319 & 0.9081 \\
27 & 0.9416 & 0.9451 & 0.9128 \\
28 & 0.9412 & 0.9412 & 0.9069 \\
29 & 0.9412 & -      & 0.9077 \\
30 & 0.9408 & -      & 0.9051 \\
31 & 0.9400 & -      & 0.9042 \\
32 & 0.9381 & -      & - \\
\bottomrule
\end{tabular}
\label{tab:llama_qwen_mistral}
\end{table}

\begin{table*}[ht]
\caption{Three-model fusion classification performance across SST-2, MR, R8, and R52 datasets.}
\centering
\begin{tabular}{lcccc}
\toprule
\textbf{} & \textbf{SST-2} & \textbf{MR} & \textbf{R8} & \textbf{R52} \\
\midrule
llama2, mistral, falcon3       & 0.9461 & 0.9588 & 0.9708 & 0.9354 \\
llama2, mistral, nv\_embed     & 0.9599 & 0.9701 & 0.9822 & 0.9601 \\
llama2, mistral, e5            & 0.9564 & 0.9633 & 0.9758 & 0.9579 \\
llama2, falcon3, nv\_embed     & 0.9564 & 0.9698 & 0.9831 & 0.9599 \\
llama2, falcon3, e5            & 0.9599 & 0.9639 & 0.9804 & 0.9591 \\
llama2, nv\_embed, e5          & 0.9610 & 0.9710 & 0.9822 & 0.9611 \\
llama2, qwen2.5, nv\_embed     & 0.9563 & 0.9706 & 0.9831 & 0.9622 \\
mistral, falcon3, nv\_embed    & 0.9599 & 0.9702 & 0.9813 & 0.9591 \\
mistral, falcon3, e5           & 0.9484 & 0.9628 & 0.9749 & 0.9540 \\
mistral, nv\_embed, e5         & 0.9610 & 0.9706 & 0.9822 & 0.9618 \\
falcon3, nv\_embed, e5         & 0.9599 & 0.9712 & 0.9831 & 0.9603 \\
qwen2.5, gemma2, nv\_embed     & 0.9610 & 0.9643 & 0.9840 & 0.9603 \\
\bottomrule
\end{tabular}
\label{tab:model_combo_accuracy}
\end{table*}

\begin{table*}[ht]
\caption{. Four-model and All-model fusion classification performance across SST-2, MR, R8, and R52 datasets.}
\centering
\begin{tabular}{lcccc}
\toprule
\textbf{} & \textbf{SST-2} & \textbf{MR} & \textbf{R8} & \textbf{R52} \\
\midrule
\multicolumn{5}{l}{\textbf{Four Models}} \\
llama2, mistral, falcon3, nv\_embed & 0.9587 & 0.9702 & 0.9813 & 0.9595 \\
llama2, mistral, falcon3, e5        & 0.9576 & 0.9636 & 0.9744 & 0.9579 \\
llama2, mistral, nv\_embed, e5      & 0.9622 & 0.9706 & 0.9808 & 0.9626 \\
llama2, qwen, nv\_embed, e5         & 0.9610 & 0.9709 & 0.9845 & 0.9638 \\
llama2, falcon3, nv\_embed, e5      & 0.9610 & 0.9712 & 0.9831 & 0.9628 \\
mistral, falcon3, nv\_embed, e5     & 0.9610 & 0.9704 & 0.9813 & 0.9618 \\
\midrule
\multicolumn{5}{l}{\textbf{All Models}} \\
All                                  & 0.9610 & 0.9719 & 0.9822 & 0.9611 \\
\bottomrule
\end{tabular}
\label{tab:four_all_model_combos}
\end{table*}

\section{Fusion-based classification results with more than three decoder-based models}

Interestingly, fusing embeddings from more than three models often resulted in lower accuracy compared to the optimal fusion of two complementary models. This observation suggests that simply increasing the number of models in the fusion does not guarantee better performance, and may even introduce redundant or conflicting information that leads to representational noise.

However, it is noteworthy that multi-model fusion still demonstrated stable and robust performance on average, indicating that while the accuracy may not always improve, the representation becomes more resilient across tasks. This may be particularly beneficial in scenarios where task-specific model selection is not feasible, or when general-purpose robustness is preferred over task-specific tuning.

An exception to this trend was observed on the R52 dataset, where the highest accuracy was achieved by concatenating embeddings from four different models. Given the larger number of classes and higher semantic diversity in R52, it is plausible that aggregating multiple embedding spaces contributed to a richer and more discriminative representation. This highlights the potential of multi-model fusion strategies in complex classification tasks with fine-grained label sets.
These findings underscore the importance of not only the number of models, but also the method of fusion and task characteristics, in determining the effectiveness of embedding combination strategies.


\begin{thebibliography}{36}
\expandafter\ifx\csname natexlab\endcsname\relax\def\natexlab#1{#1}\fi

\bibitem[{Agarap(2019)}]{agarap2019deeplearningusingrectified}
Abien~Fred Agarap. 2019.
\newblock \href {http://arxiv.org/abs/1803.08375} {Deep learning using rectified linear units (relu)}.

\bibitem[{Almazrouei et~al.(2023)Almazrouei, Alobeidli, Alshamsi, Cappelli, Cojocaru, Debbah, Étienne Goffinet, Hesslow, Launay, Malartic, Mazzotta, Noune, Pannier, and Penedo}]{almazrouei2023falconseriesopenlanguage}
Ebtesam Almazrouei, Hamza Alobeidli, Abdulaziz Alshamsi, Alessandro Cappelli, Ruxandra Cojocaru, Mérouane Debbah, Étienne Goffinet, Daniel Hesslow, Julien Launay and Quentin Malartic et~al. 2023.
\newblock \href {http://arxiv.org/abs/2311.16867} {The falcon series of open language models}.

\bibitem[{Blandfort et~al.(2019)Blandfort, Karayil, Raue, Hees, and Dengel}]{blandfort2019fusionstrategieslearninguser}
Philipp Blandfort, Tushar Karayil, Federico Raue, Jörn Hees and Andreas Dengel. 2019.
\newblock \href {http://arxiv.org/abs/1901.02322} {Fusion strategies for learning user embeddings with neural networks}.

\bibitem[{Brown et~al.(2020)Brown, Mann, Ryder, Subbiah, Kaplan, Dhariwal, Neelakantan, Shyam, Sastry, Askell, Agarwal, Herbert-Voss, Krueger, Henighan, Child, Ramesh, Ziegler, Wu, Winter, Hesse, Chen, Sigler, Litwin, Gray, Chess, Clark, Berner, McCandlish, Radford, Sutskever, and Amodei}]{brown2020languagemodelsfewshotlearners}
Tom~B. Brown, Benjamin Mann, Nick Ryder, Melanie Subbiah, Jared Kaplan, Prafulla Dhariwal, Arvind Neelakantan, Pranav Shyam, Girish Sastry and Amanda Askell et~al. 2020.
\newblock \href {http://arxiv.org/abs/2005.14165} {Language models are few-shot learners}.

\bibitem[{Cao et~al.(2024)Cao, Cai, Zhang, Zou, and Lam}]{cao2024worstpromptperformancelarge}
Bowen Cao, Deng Cai, Zhisong Zhang, Yuexian Zou and Wai Lam. 2024.
\newblock \href {http://arxiv.org/abs/2406.10248} {On the worst prompt performance of large language models}.

\bibitem[{Devlin et~al.(2019)Devlin, Chang, Lee, and Toutanova}]{devlin2019bertpretrainingdeepbidirectional}
Jacob Devlin, Ming-Wei Chang, Kenton Lee and Kristina Toutanova. 2019.
\newblock \href {http://arxiv.org/abs/1810.04805} {Bert: Pre-training of deep bidirectional transformers for language understanding}.

\bibitem[{Gardias et~al.(2020)Gardias, Arthur, and Sun}]{gardias2020enhancedresidualnetworkscontextbased}
Przemek Gardias, Eric Arthur and Huaming Sun. 2020.
\newblock \href {http://arxiv.org/abs/2005.06723} {Enhanced residual networks for context-based image outpainting}.

\bibitem[{Gretz et~al.(2023)Gretz, Halfon, Shnayderman, Toledo-Ronen, Spector, Dankin, Katsis, Arviv, Katz, Slonim, and Ein-Dor}]{gretz-etal-2023-zero}
Shai Gretz, Alon Halfon, Ilya Shnayderman, Orith Toledo-Ronen, Artem Spector, Lena Dankin, Yannis Katsis, Ofir Arviv, Yoav Katz and Noam Slonim et~al. 2023.
\newblock \href {https://doi.org/10.18653/v1/2023.findings-emnlp.647} {Zero-shot topical text classification with {LLM}s - an experimental study}.
\newblock In \emph{Findings of the Association for Computational Linguistics: EMNLP 2023}, pages 9647--9676, Singapore. Association for Computational Linguistics.

\bibitem[{Han et~al.(2024)Han, Nguyen, Harris, Ho, and Saria}]{han2024fusemoemixtureofexpertstransformersfleximodal}
Xing Han, Huy Nguyen, Carl Harris, Nhat Ho and Suchi Saria. 2024.
\newblock \href {http://arxiv.org/abs/2402.03226} {Fusemoe: Mixture-of-experts transformers for fleximodal fusion}.

\bibitem[{He et~al.(2024)He, Rungta, Koleczek, Sekhon, Wang, and Hasan}]{he2024doespromptformattingimpact}
Jia He, Mukund Rungta, David Koleczek, Arshdeep Sekhon, Franklin~X Wang and Sadid Hasan. 2024.
\newblock \href {http://arxiv.org/abs/2411.10541} {Does prompt formatting have any impact on llm performance?}

\bibitem[{Jiang et~al.(2023)Jiang, Sablayrolles, Mensch, Bamford, Chaplot, de~las Casas, Bressand, Lengyel, Lample, Saulnier, Lavaud, Lachaux, Stock, Scao, Lavril, Wang, Lacroix, and Sayed}]{jiang2023mistral7b}
Albert~Q. Jiang, Alexandre Sablayrolles, Arthur Mensch, Chris Bamford, Devendra~Singh Chaplot, Diego de~las Casas, Florian Bressand, Gianna Lengyel, Guillaume Lample and Lucile Saulnier et~al. 2023.
\newblock \href {http://arxiv.org/abs/2310.06825} {Mistral 7b}.

\bibitem[{Kaushik et~al.(2024)Kaushik, P, and Ratha}]{kaushik2024enhancingauthorshipattributionembedding}
Arjun~Ramesh Kaushik, Sunil Rufus~R P and Nalini Ratha. 2024.
\newblock \href {http://arxiv.org/abs/2411.00411} {Enhancing authorship attribution through embedding fusion: A novel approach with masked and encoder-decoder language models}.

\bibitem[{Ko et~al.(2024)Ko, Parkinson, and Wang}]{Ko2024.08.24.609531}
Young~Su Ko, Jonathan Parkinson and Wei Wang. 2024.
\newblock \href {https://doi.org/10.1101/2024.08.24.609531} {Benchmarking text-integrated protein language model embeddings and embedding fusion on diverse downstream tasks}.
\newblock \emph{bioRxiv}.

\bibitem[{Kojima et~al.(2023)Kojima, Gu, Reid, Matsuo, and Iwasawa}]{kojima2023largelanguagemodelszeroshot}
Takeshi Kojima, Shixiang~Shane Gu, Machel Reid, Yutaka Matsuo and Yusuke Iwasawa. 2023.
\newblock \href {http://arxiv.org/abs/2205.11916} {Large language models are zero-shot reasoners}.

\bibitem[{Kumar et~al.(2020)Kumar, Kumar, and Mishra}]{kumar2020quarcquaternionmultimodalfusion}
Deepak Kumar, Nalin Kumar and Subhankar Mishra. 2020.
\newblock \href {http://arxiv.org/abs/2012.08312} {Quarc: Quaternion multi-modal fusion architecture for hate speech classification}.

\bibitem[{Lee et~al.(2025)Lee, Roy, Xu, Raiman, Shoeybi, Catanzaro, and Ping}]{lee2025nvembedimprovedtechniquestraining}
Chankyu Lee, Rajarshi Roy, Mengyao Xu, Jonathan Raiman, Mohammad Shoeybi, Bryan Catanzaro and Wei Ping. 2025.
\newblock \href {http://arxiv.org/abs/2405.17428} {Nv-embed: Improved techniques for training llms as generalist embedding models}.

\bibitem[{Liu et~al.(2024)Liu, Zhang, Zhao, Ju, and Yang}]{liu2024llmembedrethinkinglightweightllms}
Chun Liu, Hongguang Zhang, Kainan Zhao, Xinghai Ju and Lin Yang. 2024.
\newblock \href {http://arxiv.org/abs/2406.03725} {Llmembed: Rethinking lightweight llm's genuine function in text classification}.

\bibitem[{Liu et~al.(2019)Liu, Ott, Goyal, Du, Joshi, Chen, Levy, Lewis, Zettlemoyer, and Stoyanov}]{liu2019robertarobustlyoptimizedbert}
Yinhan Liu, Myle Ott, Naman Goyal, Jingfei Du, Mandar Joshi, Danqi Chen, Omer Levy, Mike Lewis, Luke Zettlemoyer and Veselin Stoyanov. 2019.
\newblock \href {http://arxiv.org/abs/1907.11692} {Roberta: A robustly optimized bert pretraining approach}.

\bibitem[{Maas et~al.(2011)Maas, Daly, Pham, Huang, Ng, and Potts}]{maas-etal-2011-learning}
Andrew~L. Maas, Raymond~E. Daly, Peter~T. Pham, Dan Huang, Andrew~Y. Ng and Christopher Potts. 2011.
\newblock \href {https://aclanthology.org/P11-1015/} {Learning word vectors for sentiment analysis}.
\newblock In \emph{Proceedings of the 49th Annual Meeting of the Association for Computational Linguistics: Human Language Technologies}, pages 142--150, Portland, Oregon, USA. Association for Computational Linguistics.

\bibitem[{Pang et~al.(2002)Pang, Lee, and Vaithyanathan}]{pang2002thumbsupsentimentclassification}
Bo~Pang, Lillian Lee and Shivakumar Vaithyanathan. 2002.
\newblock \href {http://arxiv.org/abs/cs/0205070} {Thumbs up? sentiment classification using machine learning techniques}.

\bibitem[{Qwen et~al.(2025)Qwen, :, Yang, Yang, Zhang, Hui, Zheng, Yu, Li, Liu, Huang, Wei, Lin, Yang, Tu, Zhang, Yang, Yang, Zhou, Lin, Dang, Lu, Bao, Yang, Yu, Li, Xue, Zhang, Zhu, Men, Lin, Li, Tang, Xia, Ren, Ren, Fan, Su, Zhang, Wan, Liu, Cui, Zhang, and Qiu}]{qwen2025qwen25technicalreport}
Qwen, :, An~Yang, Baosong Yang, Beichen Zhang, Binyuan Hui, Bo~Zheng, Bowen Yu, Chengyuan Li and Dayiheng Liu et~al. 2025.
\newblock \href {http://arxiv.org/abs/2412.15115} {Qwen2.5 technical report}.

\bibitem[{Sarkar et~al.(2023)Sarkar, Feng, and Karmaker~Santu}]{sarkar-etal-2023-zero}
Souvika Sarkar, Dongji Feng and Shubhra~Kanti Karmaker~Santu. 2023.
\newblock \href {https://doi.org/10.18653/v1/2023.emnlp-main.1008} {Zero-shot multi-label topic inference with sentence encoders and {LLM}s}.
\newblock In \emph{Proceedings of the 2023 Conference on Empirical Methods in Natural Language Processing}, pages 16218--16233, Singapore. Association for Computational Linguistics.

\bibitem[{Shinnou et~al.(2018)Shinnou, Zhao, and Komiya}]{shinnou-etal-2018-domain}
Hiroyuki Shinnou, Xinyu Zhao and Kanako Komiya. 2018.
\newblock \href {https://aclanthology.org/Y18-1068/} {Domain adaptation using a combination of multiple embeddings for sentiment analysis}.
\newblock In \emph{Proceedings of the 32nd Pacific Asia Conference on Language, Information and Computation}, Hong Kong. Association for Computational Linguistics.

\bibitem[{Skean et~al.(2024)Skean, Arefin, LeCun, and Shwartz-Ziv}]{skean2024doesrepresentationmatterexploring}
Oscar Skean, Md~Rifat Arefin, Yann LeCun and Ravid Shwartz-Ziv. 2024.
\newblock \href {http://arxiv.org/abs/2412.09563} {Does representation matter? exploring intermediate layers in large language models}.

\bibitem[{Socher et~al.(2013)Socher, Perelygin, Wu, Chuang, Manning, Ng, and Potts}]{socher-etal-2013-recursive}
Richard Socher, Alex Perelygin, Jean Wu, Jason Chuang, Christopher~D. Manning, Andrew Ng and Christopher Potts. 2013.
\newblock \href {https://aclanthology.org/D13-1170/} {Recursive deep models for semantic compositionality over a sentiment treebank}.
\newblock In \emph{Proceedings of the 2013 Conference on Empirical Methods in Natural Language Processing}, pages 1631--1642, Seattle, Washington, USA. Association for Computational Linguistics.

\bibitem[{Sun et~al.(2023)Sun, Li, Li, Wu, Guo, Zhang, and Wang}]{sun2023textclassificationlargelanguage}
Xiaofei Sun, Xiaoya Li, Jiwei Li, Fei Wu, Shangwei Guo, Tianwei Zhang and Guoyin Wang. 2023.
\newblock \href {http://arxiv.org/abs/2305.08377} {Text classification via large language models}.

\bibitem[{Tao et~al.(2024)Tao, Shen, Gao, Zhang, Li, Tao, and Ma}]{tao2024llmseffectiveembeddingmodels}
Chongyang Tao, Tao Shen, Shen Gao, Junshuo Zhang, Zhen Li, Zhengwei Tao and Shuai Ma. 2024.
\newblock \href {http://arxiv.org/abs/2412.12591} {Llms are also effective embedding models: An in-depth overview}.

\bibitem[{Team et~al.(2024)Team, Riviere, Pathak, Sessa, Hardin, Bhupatiraju, Hussenot, Mesnard, Shahriari, Ramé, Ferret, Liu, Tafti, Friesen, Casbon, Ramos, Kumar, Lan, Jerome, Tsitsulin, Vieillard, Stanczyk, Girgin, Momchev, Hoffman, Thakoor, Grill, Neyshabur, Bachem, Walton, Severyn, Parrish, Ahmad, Hutchison, Abdagic, Carl, Shen, Brock, Coenen, Laforge, Paterson, Bastian, Piot, Wu, Royal, Chen, Kumar, Perry, Welty, Choquette-Choo, Sinopalnikov, Weinberger, Vijaykumar, Rogozińska, Herbison, Bandy, Wang, Noland, Moreira, Senter, Eltyshev, Visin, Rasskin, Wei, Cameron, Martins, Hashemi, Klimczak-Plucińska, Batra, Dhand, Nardini, Mein, Zhou, Svensson, Stanway, Chan, Zhou, Carrasqueira, Iljazi, Becker, Fernandez, van Amersfoort, Gordon, Lipschultz, Newlan, yeong Ji, Mohamed, Badola, Black, Millican, McDonell, Nguyen, Sodhia, Greene, Sjoesund, Usui, Sifre, Heuermann, Lago, McNealus, Soares, Kilpatrick, Dixon, Martins, Reid, Singh, Iverson, Görner, Velloso, Wirth, Davidow, Miller, Rahtz, Watson, Risdal,
  Kazemi, Moynihan, Zhang, Kahng, Park, Rahman, Khatwani, Dao, Bardoliwalla, Devanathan, Dumai, Chauhan, Wahltinez, Botarda, Barnes, Barham, Michel, Jin, Georgiev, Culliton, Kuppala, Comanescu, Merhej, Jana, Rokni, Agarwal, Mullins, Saadat, Carthy, Cogan, Perrin, Arnold, Krause, Dai, Garg, Sheth, Ronstrom, Chan, Jordan, Yu, Eccles, Hennigan, Kocisky, Doshi, Jain, Yadav, Meshram, Dharmadhikari, Barkley, Wei, Ye, Han, Kwon, Xu, Shen, Gong, Wei, Cotruta, Kirk, Rao, Giang, Peran, Warkentin, Collins, Barral, Ghahramani, Hadsell, Sculley, Banks, Dragan, Petrov, Vinyals, Dean, Hassabis, Kavukcuoglu, Farabet, Buchatskaya, Borgeaud, Fiedel, Joulin, Kenealy, Dadashi, and Andreev}]{gemmateam2024gemma2improvingopen}
Gemma Team, Morgane Riviere, Shreya Pathak, Pier~Giuseppe Sessa, Cassidy Hardin, Surya Bhupatiraju, Léonard Hussenot, Thomas Mesnard, Bobak Shahriari and Alexandre Ramé et~al. 2024.
\newblock \href {http://arxiv.org/abs/2408.00118} {Gemma 2: Improving open language models at a practical size}.

\bibitem[{Touvron et~al.(2023{\natexlab{a}})Touvron, Lavril, Izacard, Martinet, Lachaux, Lacroix, Rozière, Goyal, Hambro, Azhar, Rodriguez, Joulin, Grave, and Lample}]{touvron2023llamaopenefficientfoundation}
Hugo Touvron, Thibaut Lavril, Gautier Izacard, Xavier Martinet, Marie-Anne Lachaux, Timothée Lacroix, Baptiste Rozière, Naman Goyal, Eric Hambro and Faisal Azhar et~al. 2023{\natexlab{a}}.
\newblock \href {http://arxiv.org/abs/2302.13971} {Llama: Open and efficient foundation language models}.

\bibitem[{Touvron et~al.(2023{\natexlab{b}})Touvron, Martin, Stone, Albert, Almahairi, Babaei, Bashlykov, Batra, Bhargava, Bhosale, Bikel, Blecher, Ferrer, Chen, Cucurull, Esiobu, Fernandes, Fu, Fu, Fuller, Gao, Goswami, Goyal, Hartshorn, Hosseini, Hou, Inan, Kardas, Kerkez, Khabsa, Kloumann, Korenev, Koura, Lachaux, Lavril, Lee, Liskovich, Lu, Mao, Martinet, Mihaylov, Mishra, Molybog, Nie, Poulton, Reizenstein, Rungta, Saladi, Schelten, Silva, Smith, Subramanian, Tan, Tang, Taylor, Williams, Kuan, Xu, Yan, Zarov, Zhang, Fan, Kambadur, Narang, Rodriguez, Stojnic, Edunov, and Scialom}]{touvron2023llama2openfoundation}
Hugo Touvron, Louis Martin, Kevin Stone, Peter Albert, Amjad Almahairi, Yasmine Babaei, Nikolay Bashlykov, Soumya Batra, Prajjwal Bhargava and Shruti Bhosale et~al. 2023{\natexlab{b}}.
\newblock \href {http://arxiv.org/abs/2307.09288} {Llama 2: Open foundation and fine-tuned chat models}.

\bibitem[{Wang et~al.(2024)Wang, Yang, Huang, Jiao, Yang, Jiang, Majumder, and Wei}]{wang2024textembeddingsweaklysupervisedcontrastive}
Liang Wang, Nan Yang, Xiaolong Huang, Binxing Jiao, Linjun Yang, Daxin Jiang, Rangan Majumder and Furu Wei. 2024.
\newblock \href {http://arxiv.org/abs/2212.03533} {Text embeddings by weakly-supervised contrastive pre-training}.

\bibitem[{Wei et~al.(2023)Wei, Wang, Schuurmans, Bosma, Ichter, Xia, Chi, Le, and Zhou}]{wei2023chainofthoughtpromptingelicitsreasoning}
Jason Wei, Xuezhi Wang, Dale Schuurmans, Maarten Bosma, Brian Ichter, Fei Xia, Ed~Chi, Quoc Le and Denny Zhou. 2023.
\newblock \href {http://arxiv.org/abs/2201.11903} {Chain-of-thought prompting elicits reasoning in large language models}.

\bibitem[{Youngmin et~al.(2024)Youngmin, Andrew, Duoduo, and Stephen}]{youngmin2024rolemodelarchitecturescale}
Lee Youngmin, Lang S. I.~D. Andrew, Cai Duoduo and Wheat~R. Stephen. 2024.
\newblock \href {http://arxiv.org/abs/2405.00949} {The role of model architecture and scale in predicting molecular properties: Insights from fine-tuning roberta, bart, and llama}.

\bibitem[{Zhang et~al.(2022)Zhang, Zhang, Huang, and Yu}]{zhang-etal-2022-prompt-based}
Haoxing Zhang, Xiaofeng Zhang, Haibo Huang and Lei Yu. 2022.
\newblock \href {https://doi.org/10.18653/v1/2022.emnlp-main.87} {Prompt-based meta-learning for few-shot text classification}.
\newblock In \emph{Proceedings of the 2022 Conference on Empirical Methods in Natural Language Processing}, pages 1342--1357, Abu Dhabi, United Arab Emirates. Association for Computational Linguistics.

\bibitem[{Zhang et~al.(2003)Zhang, Jin, Yang, and Hauptmann}]{inproceedings}
Jian Zhang, Rong Jin, Yiming Yang and Alexander Hauptmann. 2003.
\newblock Modified logistic regression: An approximation to svm and its applications in large-scale text categorization.
\newblock volume~2, pages 888--895.

\bibitem[{Zhang et~al.(2024)Zhang, Dong, and Kawaguchi}]{zhang2024investigatinglayerimportancelarge}
Yang Zhang, Yanfei Dong and Kenji Kawaguchi. 2024.
\newblock \href {http://arxiv.org/abs/2409.14381} {Investigating layer importance in large language models}.

\end{thebibliography}
\end{document}